\documentclass[a4paper,fleqn]{cas-dc}

\usepackage[authoryear]{natbib}
\RequirePackage{color}
\RequirePackage{algorithm}
\RequirePackage{algorithmic}
\usepackage{microtype}
\usepackage{graphicx}
\usepackage{subfigure}
\usepackage{amssymb}
\usepackage{booktabs}
\usepackage{hyperref}
\usepackage{amsmath}

\usepackage{xcolor,soul}
\hypersetup{
colorlinks=true,    % false: boxed links; true: colored links
linkcolor=blue, % color of internal links
citecolor=blue,   % color of links to bibliography
filecolor=blue, % color of file links
urlcolor=blue   % color of external links
}

\def\tsc#1{\csdef{#1}{\textsc{\lowercase{#1}}\xspace}}
\tsc{WGM}
\tsc{QE}
\tsc{EP}
\tsc{PMS}
\tsc{BEC}
\tsc{DE}

\begin{document}
\let\WriteBookmarks\relax
\def\floatpagepagefraction{1}
\def\textpagefraction{.001}
\shorttitle{Neural Ordinary Differential Equations for Dynamics Modeling in Continuous Control}
\shortauthors{V. M. Martinez Alvarez et~al.}

\title [mode = title]{DyNODE: Neural Ordinary Differential Equations for Dynamics Modeling in Continuous Control}                      

\author[1]{Victor M. Martinez Alvarez}
\cormark[1]
% \fnmark[1]
\ead{martinez@rist.ro}

\address[1]{Romanian Institute of Science and Technology, Cluj-Napoca, Romania}

\author[1]{Rare\textcommabelow{s} Ro\textcommabelow{s}ca}

\author[1]{Cristian G. F\u{a}lcu\textcommabelow{t}escu}

\begin{abstract}
We present a novel approach (DyNODE) that captures the underlying dynamics of a system by incorporating control in a neural ordinary differential equation framework. We conduct a systematic evaluation and comparison of our method and standard neural network architectures for dynamics modeling. Our results indicate that a simple DyNODE architecture when combined with an actor-critic reinforcement learning (RL) algorithm that uses model predictions to improve the critic's target values, outperforms canonical neural networks, both in sample efficiency and predictive performance across a diverse range of continuous tasks that are frequently used to benchmark RL algorithms. This approach provides a new avenue for the development of models that are more suited to learn the evolution of dynamical systems, particularly useful in the context of model-based reinforcement learning.  To
assist related work, we have made code available
at \href{ https://github.com/vmartinezalvarez/DyNODE}{ https://github.com/vmartinezalvarez/DyNODE}.
\end{abstract}


\maketitle

\section{Introduction}
\label{intro}

Deep reinforcement learning (RL) is currently one of the most successful applications of deep learning, with use in a wide range of continuous control and decision making tasks, ranging from robotics to video games~\citep{Lili, mnih, silver2016mastering}. In the last few years, the search for new reinforcement learning algorithms has turned to model-based reinforcement learning (MBRL)~\citep{MBMF}. In this context, learning dynamics models that are accurate enough for planning is a challenging and important problem~\citep{moore}. Previously, advancements in deep RL have generally revolved around model-free approaches, which do not attempt to learn a dynamics model, but rather use direct interaction with the environment to learn a policy or a value function~\citep{schulman, levi}. However, while these algorithms are generally better in terms of asymptotic performance and can easily be extended to high dimensional problems, their success in terms of performance comes at the cost of being data-expensive, since they require a large number of interactions with the environment. This can be impractical in robotic systems and thus largely limits their applications to simulated domains. On the other hand, by learning a model of the environment and then using it as a substitute of the real system, MBRL algorithms have the potential of being significantly more sample efficient than their model-free counterparts~\citep{MBMF, want2019}. 

Model-based RL approaches usually struggle to match the asymptotic performance of model-free algorithms ~\citep{benchmark}. In particular, policy improvement takes into consideration a learned model, which might not accurately capture the true dynamics of the system. Thus, inaccurate predictions are made with high-confidence and changes in policy parameters are (falsely) predicted to have a significant effect on the expected return. As a result, such policies will exploit these deficiencies and get stuck at suboptimal behaviors. In cases where few samples are available to the agent to learn a model, model-bias – a problem that emerges when selecting with certainty only one dynamics model from a large possible collection - gets exacerbated and learning a faithful representation of the environment becomes even more challenging~\citep{gal2016improving}. Furthermore, due to the curse of dimensionality, in high dimensional environments even very small errors tend to compound and propagate, especially for long planning horizons.

Recent advancements in the field have managed to reduce the effects of model-bias through the use of probabilistic models~\citep{trials} and ensembles~\citep{ME-TRPO}, which incorporate model uncertainty into planning and decision making and help learn policies that are more robust to model errors.  In this context, high-capacity function approximators are usually considered one of the default starting point for dynamics modeling in MBRL. However, neural networks are notoriously data hungry and prone to overfitting when data is scarce. Furthermore, since exploration is coupled with the policy \citep{schmid}, algorithms with learned dynamics that are stuck at suboptimal behaviors do not increase the performance when collecting more data, an issue known as dynamics bottleneck \citep{benchmark}.  

On the other hand, many of the environments used to benchmark continuous control RL algorithms are modeled by differential equations, i.e., the state evolves according to a system of ordinary differential equations (ODE), which describes the underlying physical laws that govern the environment, namely the transition function~\citep{tassa2018deepmind}. We aim to exploit the similarities between the way the states evolve in these systems and the Neural Ordinary Differential Equations (NODEs)~\citep{NODE} paradigm, and establish a natural connection between them to learn a world model. 

Inspired by this physical connection, we propose a new method that is better fit to distill the dependencies of systems whose evolution is dictated by differential equations. The main contribution of our work is the proposal of a novel mechanism (DyNODE) to learn dynamics models in continuous control by integrating actions in the NODE~\citep{NODE} framework while propagating the state value errors. Also, we demonstrate that such a model can reduce the effect of error compounding from model predictions across the imagination length, for both low and high dimensional environments. Moreover, we show that DyNODE, when combined with a model-free reinforcement learning algorithm that uses model predictions to improve value estimation, outperforms canonical neural networks both in sample efficiency and predictive performance across a diverse range of continuous tasks that are frequently used to benchmark RL algorithms.

\section{Related Work}

\textbf{Probabilistic models:} Probabilistic models have been used in MBRL algorithms that perform a policy search with backpropagation through time for many years, with
PILCO ~\citep{PILCO} being a prominent example. Instead of using a model to simulate the interaction of the agent with the environment and gather imagined data, these methods compute an analytic gradient of the objective with respect to the parameters of the policy and update the policy accordingly. The dynamics models in these methods are probabilistic, with function approximators being Gaussian processes~\citep{PILCO}, time-varying Gaussian-linear processes~\citep{GPS1, GPS2, GPS3} or a mixture of Gaussian processes~\citep{mixture}. The dynamics model in the SVG algorithm~\citep{SVG} is also probabilistic, but uses observations from the real environment, rather than from the simulated one. These probabilistic model approaches are data-efficient but often struggle to scale to high-dimensional environments and nonlinear dynamics~\citep{MBMF, benchmark}, suffering from the problem of vanishing or exploding gradient. The results presented by Wang et al.~\citep{benchmark} show that these methods are outperformed by Dyna-style or Shooting algorithms in all but two out of 18 OpenAI Gym \citep{open} environments.

\textbf{Ensembles:} Model ensembles have been successfully used in Dyna-style algorithms~\citep{dyna} to improve their performance. Instead of using a single model to capture the dynamics of the system, such methods~\citep{MB-MPO, ME-TRPO, SLBO} use an ensemble to mitigate the effects of model imperfections, while also maintaining uncertainty in the decision-making process. MB-MPO~\citep{MB-MPO}, for example, manages to reduce even more the reliance on an accurately learned model by meta-learning a policy that can adapt to any of the dynamic models in the ensemble in one gradient step. While an ensemble of networks results in a more precise simulation of the environment, the individual models that compose the ensemble use standard neural network architectures as the function approximator. Chua et al.~\citep{trials} introduce PETS, a state-of-the-art MBRL algorithm that consists of an ensemble of probabilistic neural networks and uses particle-based trajectory sampling. In~\citep{buckman2018sample}, the authors propose a stochastic ensemble value expansion (STEVE), that dynamically interpolates between model rollouts of various horizon lengths for each individual example, generalizing Model-Based Value Expansion (MVE)~\citep{feinberg2018model} that use only a single learned dynamics model to improve value estimation.

\begin{figure*}%[ht]
\centering
\vskip 0.2in
\begin{center}
\centerline{\includegraphics[width=\textwidth]{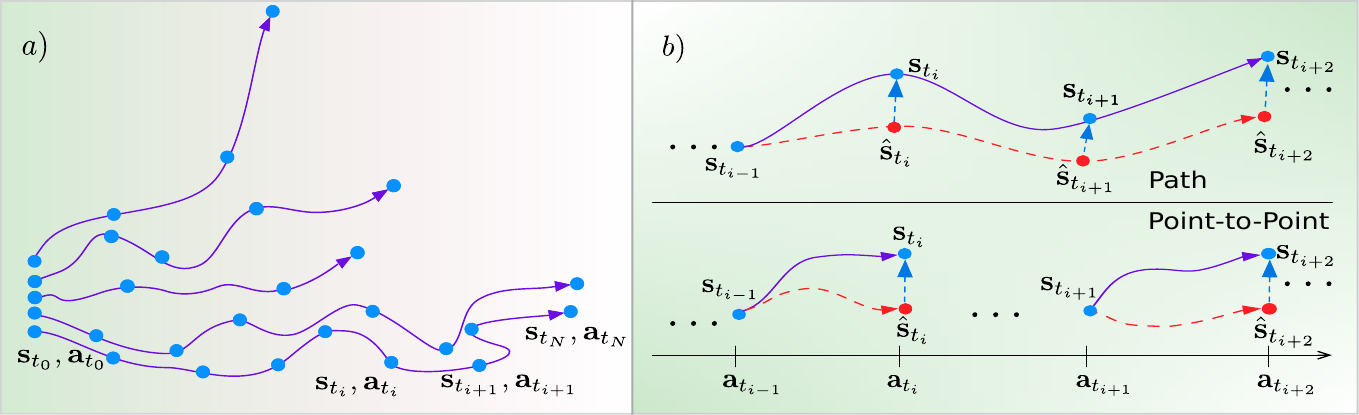}}
\caption{$a)$  Using a uniform random policy, we sample a variable number of rollouts, where each of them can have different lengths. $b)$ In the neural network baseline (down), the sample tuples contains only two consecutive states and the action, and the loss function considers the predicted state (red) for only a single step. In the DyNODE case (top), we sample entire rollouts and propagate the state value predictions through a NODE dynamics model while minimizing the prediction error across the whole horizon length.}
\label{figure_1}
\end{center}
\vskip -0.2in
\end{figure*}

\section{Background}
\subsection{Reinforcement Learning} We represent an environment as a discrete-time finite Markov decision process (MDP) $\mathcal{M}$ defined by the tuple $(\mathcal{S},\mathcal{A},f,r,\gamma,p_0,H)$. Here, $\mathcal{S}$ is the state space, $\mathcal{A}$ is the action space, $s_{t+1}\sim f(s_t,a_t)$ is the transition distribution, $r$: $\mathcal{S} \times \mathcal{A} \rightarrow \mathbb{R}$ is the reward function,  $\gamma$ is the discount factor, $p_0: \mathcal{S} \rightarrow \mathbb{R}$ represents the initial state distribution, and $H$ is the horizon of the process. The goal of reinforcement learning is to find a policy $\pi_{\phi}: \mathcal{S} \times \mathcal{A} \rightarrow \mathbb{R}$ that maximizes the expected return, i.e.,
\begin{equation}
\max_{\phi}J(\phi)=\max_{\phi}\mathbb{E}\left[\sum_{t}\gamma^{t}r(\mathbf{s}_{t},\mathbf{a}_{t})\right].
\end{equation}
Usually, RL attempts to learn the parameters $\phi$ of the policy $\pi_{\phi}(\mathbf{a}_t|\mathbf{s}_t)$ such that the expected sum of rewards is maximized under the induced trajectory distribution $\rho_{\pi}$. We can modify this objective to incorporate an entropy term, such that the policy now also aims to maximize the expected entropy $H(\pi_{\phi}(\cdotp|s_t))$ under the induced $\rho_{\pi}$. Therefore, the  maximum entropy objective becomes:
\begin{equation}
\phi^{*}=\arg\max_{\phi}\sum_{t=1}^{T}\underset{(\mathbf{s}_{t},\mathbf{a}_{t})\sim\rho_{\pi}}{\mathbb{E}}[r(\mathbf{s}_{t},\mathbf{a}_{t})+\alpha\mathcal{H}(\pi_{\phi}(\cdotp|\mathbf{s}_{t}))],
\end{equation}
where $\alpha$ is a temperature parameter that controls the trade-off between optimizing for the reward and for the entropy of the policy. This maximum entropy RL framework is used in SAC~\citep{haarnoja2018soft} to derive a soft policy iteration, alternating between policy evaluation and policy improvement. To handle continuous action spaces, SAC then extends this soft policy iteration by using parameterized function approximators to represent both the Q-function $Q_{\theta}$ (critic) and the policy $\pi_{\phi}$ (actor). The soft Q-function parameters $\theta$ are optimized to minimize the soft Bellman residual,
\begin{equation}
J_{Q}(\theta)=\underset{(\mathbf{s}_{t},\mathbf{a}_{t},r_{t},\mathbf{s}_{t+1})\sim\mathcal{D}}{\mathbb{E}}\left[\frac{1}{2}\left(Q_{\theta}(\mathbf{s}_{t},\mathbf{a}_{t})-(r_{t}+\gamma V_{\bar{\theta}}(\mathbf{s}_{t+1}))\right)^{2}\right],
\end{equation}
\begin{equation}
V_{\bar{\theta}}(\mathbf{s}_{t+1})=\underset{\mathbf{a}_{t+1}\sim\pi_{\phi}}{\mathbb{E}}\left[Q_{\bar{\theta}}(\mathbf{s}_{t+1},\mathbf{a}_{t+1})-\alpha\log\pi_{\phi}(\mathbf{a}_{t+1}|\mathbf{s}_{t+1})\right],
\end{equation}
where $D$ is the replay buffer, $\gamma$ is the discount factor, and $\bar{\theta}$ are delayed parameters. The policy parameters $\phi$ are optimized to update the policy towards the exponential of the soft Q-function,
\begin{equation}
J_{\pi}(\phi)=\underset{\mathbf{s}_{t}\sim\mathcal{D}}{\mathbb{E}}\left[\underset{\mathbf{a}_{t}\sim\pi_{\phi}}{\mathbb{E}}[\alpha\log\pi_{\phi}(\mathbf{a}_{t}|\mathbf{s}_{t})-Q_{\theta}(\mathbf{s}_{t},\mathbf{a}_{t})]\right].
\end{equation}
It has been shown that there is an increase in robustness and stability when using this stochastic, entropy maximizing RL framework~\citep{haarnoja2018soft}. Hence, by using the learned model to improve the target critic values, the learning procedure resembles that of an on-policy method while still being off-policy. 

In fact, an alternative to using model rollouts for direct training of a policy is to improve the quality of target values of the samples collected from the real environment. For this purpose, we chose  Model-Based Value Expansion (MVE)~\citep{feinberg2018model}, which showed that a learned dynamics model can be used to improve value estimation. Specifically, MVE forms temporal difference (TD) targets by combining a short term value estimate formed by unrolling the model dynamics and a long term value estimate using the learned Q function. When the model is accurate, this reduces the bias of the targets, leading to improved performance. Thus, to better determine the relationship between training on model-generated data and using model predictions to improve target values, we augment SAC with the H-step Q-target objective as follows:
\begin{equation}
    \frac{1}{H}\sum_{t=-1}^{H-1}\left(Q_{\theta}(\hat{\mathbf{s}}_{t},\hat{\mathbf{a}}_{t})-\left(\sum_{t=1}^{H-1}\gamma^{k-t}\hat{r}_{k}+\gamma^{H}Q_{\theta}(\hat{\mathbf{s}}_{H},\hat{\mathbf{a}}_{H})\right)\right)^{2}.
\end{equation}

\subsection{Neural Ordinary Differential Equations}
\label{Preliminaries}

A myriad of processes in the real world are described through the evolution of continuous hidden states, which produce observations that are in turn sequential and continuous. The behavior of these hidden states is governed by differential equations, hence any attempt to predict their future configuration, i.e., solving the initial value problem (IVP), should be based on temporal integration methods that are appropriate for solving such a system of differential equations. Neural Ordinary Differential Equations~\citep{NODE} use neural network blocks to parameterize the local derivative of the input across the interval between the input and the output. In order to compute the output, the derivative is fed to a black-box ODE solver with some numerical integration method. In particular, one block of a residual network produces the following mapping:
\begin{equation}
    \textbf{h}_{t+1} = \textbf{h}_{t} + f_{\theta}(\textbf{h}_t),
\end{equation}
where $\theta$ are the parameters of a neural network $f_{\theta}$, $\textbf{h}_{t}$ and $\textbf{h}_{t+1}$ are an input and an output of a residual block. This mapping corresponds to a one-step Euler method for numerical integration. Neural ODEs connect residual neural networks (ResNets) with differential equations and extend this idea by considering a ResNet to be a finite and discrete approximation to the solution of an ODE. In the ResNet forward update function,
\begin{equation}
    \textbf{h}_{t+1} = \textbf{h}_{t} + f(\textbf{h}_t, \theta_t),
\end{equation}
we can consider $f(\textbf{h}_t, \theta_t)$ to be the parametrization of a vector field describing the evolution of the hidden state $\textbf{h}_t\in\mathbb{R}^{D}$ at time $t$, for some parameters $\theta_t$ which usually denote the weights of a neural network. By going to the limit of very small time-steps, and treating $t$ and $\theta_t$ as independent parameters, we obtain the following ODE problem:
\begin{equation}
    \frac{d\textbf{h}(t)}{dt} = f(\textbf{h}(t), t, \theta).
\end{equation}
%%%%%%%%%%%%%%%%%%%%%%%%%%%%%%%%%%%%%%%%%%%%%%%%%%%%%%%%%%

\section{Dynamics Learning using DyNODE}

In this section, we describe the NODE based dynamics model architecture (DyNODE) which can learn the vector field associated to the transition function of the  environment,
\begin{equation}
\frac{d\textbf{s}(t)}{dt} = f(\textbf{s}(t), \textbf{a}(t),t, \theta),\label{deriva}
\end{equation}
when control is included in the form of policy-generated actions. In control problems, actions impact the evolution of the states because they usually involve applying forces on certain elements (such as joints) of the agent that interacts with the environment, thus affecting its future states. A controller takes one action, or a series of actions $\{\textbf{a}_t\}$, which ideally would lead to the specific goal-state of the environment. Generally, the actions are sampled via a policy that takes into consideration the current state. Changes in the state of the environment are dictated by an underlying transition or dynamics function $f:\mathcal{S} \times \mathcal{A} \rightarrow \mathcal{S}$ which is generally unknown. Our goal is to find a function $\hat{f_\theta}$ that can approximate the true dynamics $f$ and which can be learned from a given finite amount of data points.

If we consider the dynamics function $\hat{f_\theta}$ as the derivative of the state with respect to time as in Eq.~(\ref{deriva}), we can use an ODE solver to obtain approximate values for the next state of the environment given the current state and action by integrating over a time interval. One important assumption that this numerical evaluation must take into account is that the action remains constant along the trajectory in the state-action space, i.e., over the time interval between the current state and the next one. This is true since the action is produced by an external policy, and only then does the system evolve for one time-step. Hence, we can parametrize the derivative of the hidden state by a neural network with weights $\theta$, and use numerical methods to compute an approximation of the next state, $\hat{\textbf{s}}_{t+1}$, by integrating the following equation:
\begin{equation}
    \hat{\textbf{s}}_{t+1}=\,\textbf{s}_{t}+\int_{t}^{t+1}\hat{f}(\textbf{s}_{t}, \textbf{a}_{t}, t, \theta)dt.\label{integral}
\end{equation}
This equation can be used within a NODE architecture to learn $\hat{f_\theta}$ by sampling data-points from a replay buffer that stores transitions from past interactions with the environment. There are two main choices here: (i) sample only data points of two consecutive states and the corresponding action ($\textbf{s}_{t}, \textbf{a}_{t}, \textbf{s}_{t+1}$), or (ii) sample sequences of actions and the associated sequences of states. The first choice being the conventional way of training dynamics models, while the second one is our contribution that allows propagating the state value prediction through a NODE dynamics model for a specific horizon length.

Instead of doing prediction state to next state we can extend the DyNODE model to include a whole sequence of actions and try to predict not just the final step, but also intermediate ones Fig.~\ref{figure_1}b (top), that is, propagating state values through a neural ODE dynamics model. As we will demonstrate empirically, this allows the model to better learn the transition function, better generalize the learned dynamics to samples outside of the training set when compared to a canonical neural network that uses only next state prediction during training.

In order to predict entire trajectories, we sample rollouts of the form $(\textbf{s}_t, \textbf{A}_t, \textbf{S}_t)$ where $\textbf{s}_t$ is the initial starting state and $\textbf{A}_t=(\textbf{a}_t,\ldots,\textbf{a}_{t+H-1})$ and $\textbf{S}_t=(\textbf{s}_{t+1},\ldots,\textbf{s}_{t+H})$ are vectors containing the sequence of actions and associated next states, and $H$ is the horizon length of the action sequence. The model then uses the parametrized derivative of the hidden state in order to compute the next state~(\ref{integral}) for each subsequent time-step. The loss of this model is calculated as the mean of the absolute value of the difference between the predicted and the true states over the entire rollout,
\begin{equation}
\mathcal{L}_{\textsuperscript{Path}}=\frac{1}{\left|\mathcal{S}\right|}\sum_{i=1}^{\left|\mathcal{S}\right|}(\frac{1}{H}\sum_{h=1}^{H}\left| {s}^{i}_{t+h} - \hat{s}^{i}_{t+h} \right|), \label{path_eq}  
\end{equation}
which resembles the PILCO algorithm~\citep{PILCO}, that analytically propagates uncertain state distributions through a Gaussian process dynamics model (see algorithm~\ref{alg_path_1}). Here, upper-script denotes the $i$-th dimension of the state and the lower-script denotes the $h$-th state from the horizon length $H$.

\begin{figure*}%[ht]
\vskip 0.2in
\begin{center}
\centerline{\includegraphics[width=\textwidth]{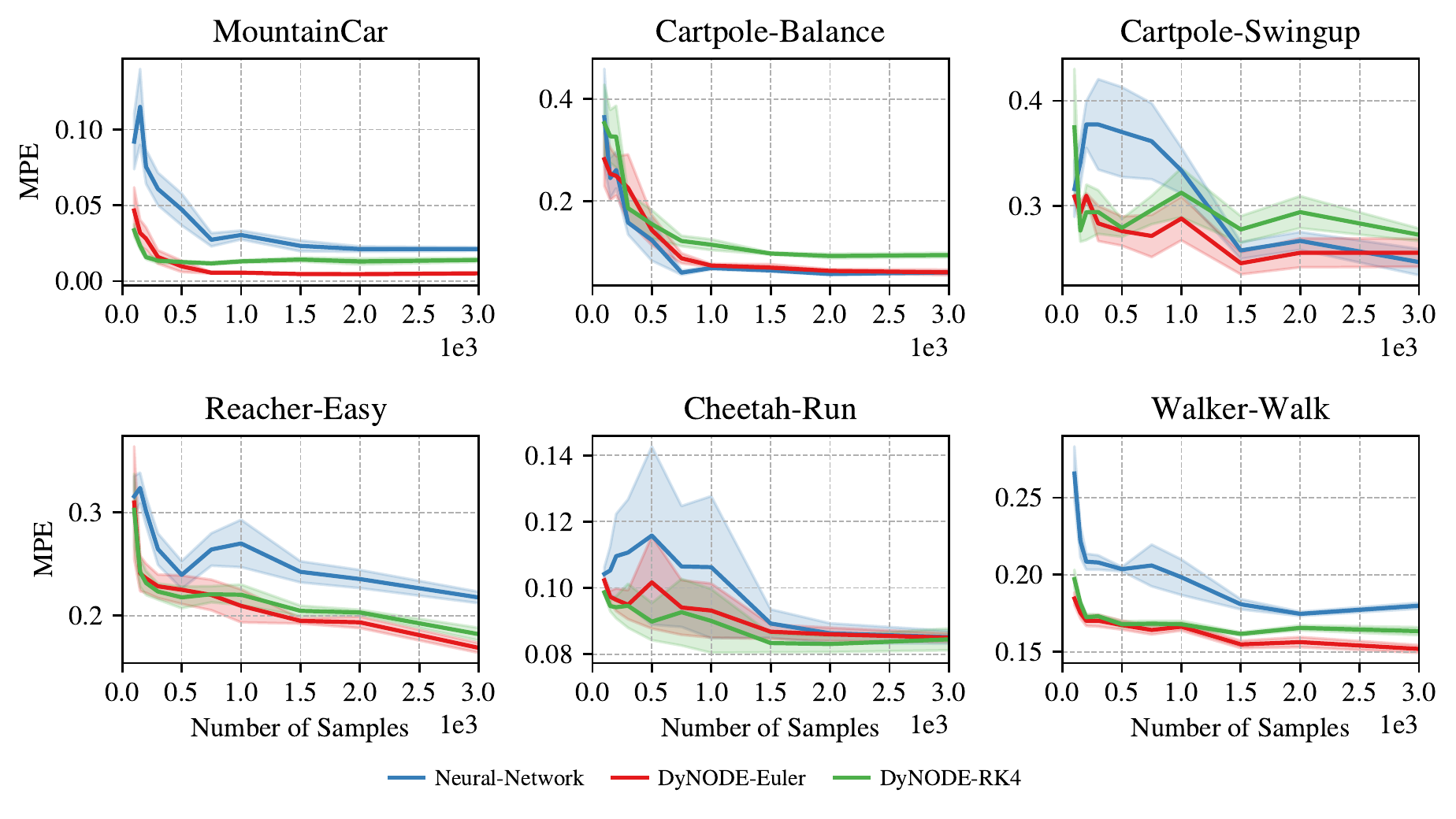}}
\caption{Mean prediction error, with standard deviation, for both DyNODE models (Euler and RK4) and the neural network baseline. Horizontal axis shows the number of environment samples used for training. The MPE is calculated for $10$ evaluation sequences of horizon length $H=200$ each, and over $5$ random seeds, collected with random policies.}
\label{figure_2}
\end{center}
\vskip -0.2in
\end{figure*}

\begin{algorithm}
\caption{DyNODE}
\label{alg_path_1}
\begin{algorithmic}[1]
\STATE Gather dataset $\mathcal{D}$ of random trajectories
\STATE Initialize dynamics model $\hat{f}_\theta$
\STATE {\bfseries Input:} evaluation steps $m$, $T=\{t_1,..., t_{m*\left|A\right|}\}$, actions-sequence length $\left|A\right|$
\STATE $\hat{\textbf{S}}=\mathrm{List()}$
\FOR{$\mathrm{iter}=1$ {\bfseries to} $\mathrm{max\_iter}$}
\FOR{$h=0$ to $\left|A\right|$}
\STATE $T'=\{t_h,..., t_{h+m}\}$
\STATE $\hat{\textbf{s}}_{t+h+1} = \mathrm{ODESolver}(\hat{f}_{\theta}, \textbf{s}_{t+h}, \textbf{a}_{t+h}, \Delta t, m)$
\STATE Append $\hat{\textbf{s}}_{t+h+1}$ to $\hat{\textbf{S}}$
\ENDFOR
\STATE train $\hat{f}_\theta$ by performing gradient descent on Eq.~(\ref{path_eq})
\ENDFOR
\end{algorithmic}
\end{algorithm}

\begin{table*}[t]
\caption{Mean-prediction error for the $6$ environments of the DyNODE model (using Euler or RK4 numerical integrator) and the neural network baseline. The $3$ large rows were chosen in order to show performance on both low and high sample regimes.}
\label{table_1}
\vskip 0.1in
\begin{center}
\begin{small}
\begin{sc}
\begin{tabular}{clccccc}
\toprule
\textbf{No. Samples} & \textbf{Environment} & \textbf{NN} & \textbf{DyNODE-Euler} & \textbf{DyNODE-RK4}\\ 
\midrule

& CartPole-Swingup       & 0.371 $\pm$ 0.065& 0.309$\pm$ 0.003& \textbf{0.293$\pm$ 0.026} \\
& MountainCar     & 0.074$\pm$ 0.011& 0.028$\pm$ 0.007& \textbf{0.015$\pm$ 0.001} \\
200 & CartPole-Balance     & 0.261$\pm$ 0.036& \textbf{0.249$\pm$ 0.038}& 0.326$\pm$ 0.061\\
& Cheetah-Run  & 0.109$\pm$ 0.015& \textbf{0.096$\pm$ 0.003}& \textbf{0.094$\pm$ 0.003} \\
& Reacher-Easy       & 0.301$\pm$ 0.014& \textbf{0.236$\pm$ 0.014}& \textbf{0.232$\pm$ 0.012} \\
& Walker-Walk      & 0.208$\pm$ 0.005& \textbf{0.170$\pm$ 0.003}& \textbf{0.172$\pm$ 0.002} \\

\midrule
& CartPole-Swingup       & 0.370 $\pm$ 0.043& \textbf{0.275$\pm$ 0.014}& \textbf{0.278$\pm$ 0.009} \\
& MountainCar     & 0.047$\pm$ 0.010& \textbf{0.010$\pm$ 0.004}& \textbf{0.012$\pm$ 0.001} \\
500 & CartPole-Balance     & \textbf{0.123$\pm$ 0.041}& 0.143$\pm$ 0.029& 0.154$\pm$ 0.028 \\
& Cheetah-Run  & 0.116$\pm$ 0.027& 0.102$\pm$ 0.014& \textbf{0.089$\pm$ 0.006} \\
& Reacher-Easy       & 0.239$\pm$ 0.013& 0.225$\pm$ 0.013& \textbf{0.218$\pm$ 0.011} \\
& Walker-Walk      & 0.204$\pm$ 0.002& \textbf{0.167$\pm$ 0.003}& \textbf{0.168$\pm$ 0.002}\\

\midrule
& CartPole-Swingup      & \textbf{0.246 $\pm$ 0.013}& 0.253$\pm$ 0.015& 0.272$\pm$ 0.006 \\
& MountainCar     & 0.021$\pm$ 0.001& \textbf{0.005$\pm$ 0.001}& 0.014$\pm$ 0.001\\
3000 & CartPole-Balance     & \textbf{0.060$\pm$ 0.003}& \textbf{0.059$\pm$ 0.005}& 0.094$\pm$ 0.005 \\
& Cheetah-Run  & \textbf{0.085$\pm$ 0.002}& \textbf{0.085$\pm$ 0.001}& \textbf{0.084$\pm$ 0.003} \\
& Reacher-Easy       & 0.218$\pm$ 0.006& \textbf{0.168$\pm$ 0.005}& 0.182$\pm$ 0.006 \\
& Walker-Walk      & 0.180$\pm$ 0.002& \textbf{0.152$\pm$ 0.003}& 0.163$\pm$ 0.003 \\
\bottomrule
\end{tabular}
\end{sc}
\end{small}
\end{center}
\vskip -0.1in
\end{table*}

\begin{figure*}%[ht]
\vskip -0.1in
\begin{center}
\centerline{\includegraphics[width=\textwidth]{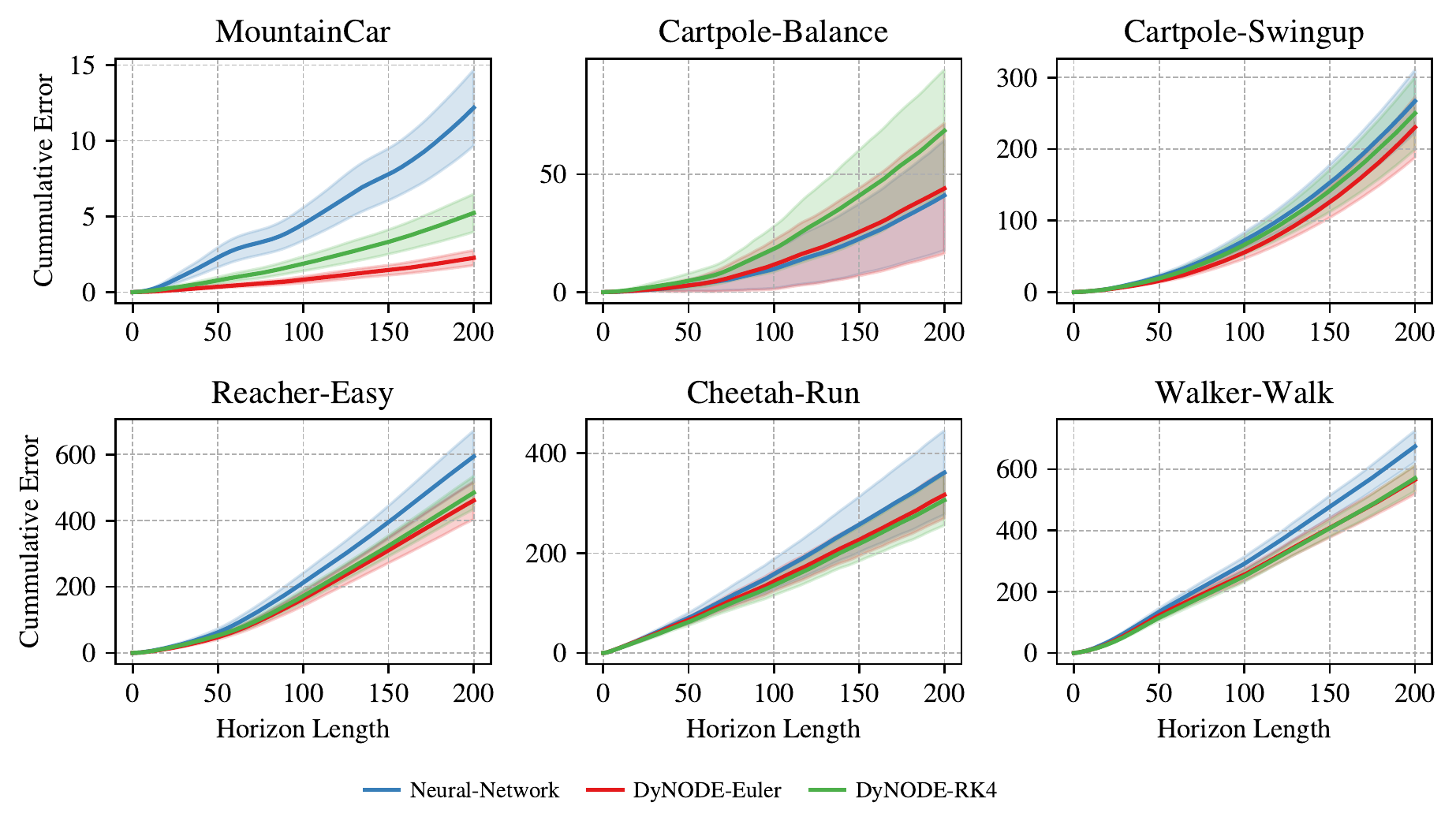}}
\caption{Cumulative prediction error for $6$ different continuous tasks.  Horizontal axis show the horizon length of the action sequences. The cumulative error is averaged over $10$ evaluation sequences and $5$ random seeds, collected with random policies.}
\label{figure_4}
\end{center}
\vskip -0.3in
\end{figure*}

\section{Results}
In order to test the DyNODE's performance, we first conducted a series of experiments on modelling continuous dynamics on six environments, five from DeepMind Control Suite~\citep{tassa2018deepmind} and one  environment (MountainCar) from OpenAI Gym~\citep{open}. The low-dimensionality of the latter allows for an easier interpretation of its dynamics, since the evolution of the entire system can be visualized in a single 2D phase space diagram. Then, we used the DyNODE algorithm coupled with SAC in a RL model-based framework that make use of model rollouts to improve the critic target values on  six DeepMind Control Suite environments~\citep{tassa2018deepmind} in order to assess its performance on different tasks. The results of all experiments are averaged across 5 random seeds.

\subsection{Random Policy Trajectories}
The samples were collected using a uniform random policy for all environments, for both training and evaluation (see Fig.~\ref{figure_1}a). We applied noise to the states for better performance~\citep{noise, want2019}. As for the evaluation, $10$ sequences of length $H=200$ steps were collected, and the same evaluation trajectories were used for all the models including the baseline. For open-loop dynamics prediction, we compare the corresponding ground truth states $(\textbf{s}_{t+1},\ldots,\textbf{s}_{t+H})$ to the dynamics model state predictions $(\hat{\textbf{s}}_{t+1},\ldots,\hat{\textbf{s}}_{t+H})$, generated by each sequence of actions of the form
$(\textbf{a}_t,\ldots,\textbf{a}_{t+H-1})$,

In Fig.~\ref{figure_2}, we show the overall dependency of the mean-prediction error (MPE) with respect to the number of samples collected from the environment. The MPE was computed over all state dimensions, across the entire length ($H$) of the sequence, and across $10$ different evaluation sequences. As the plots show, both DyNODE models (Euler and RK4) have a better overall performance than the baseline model on all but one of the environments (CartPole-Balance) on the few-samples regimes. When the number of samples is large, they match or surpass the asymptotic performance of the baseline. Moreover, the standard deviation is considerably lower in most cases for our model, indicating higher stability than the baseline on the few-samples regime. The DyNODE Euler and RK4 models, trained with an action sequence length of 20 and 7 respectively, has a significantly smaller MPE on the high-dimensional environments, especially Reacher-Easy and Walker-Walk, while exhibiting very good sample efficiency on Cheetah-Run. 

Table~\ref{table_1} presents the MPE values for all the environments for the two DyNODE models (Euler and RK4) and the neural network baseline for $3$ different sizes of the training dataset. The results showcase the good performance of the DyNODE models for both low and high-dimensional environments and both samples regimes. In Fig.~\ref{figure_4} we illustrate the variation of the cumulative prediction error as the length of the sequence of actions increases. In this experiment, DyNODE yields better generalization in 5 out of 6 tasks, which shows that this method alleviates the effect of error accumulation across the horizon length of the action sequences. The models in the figure were trained using $1000$ data-points. The DyNODE matched or outperformed the baseline neural network on both high and low sample regimes. In addition, DyNODE models performs particularly well in the high-dimensional environments, resulting in very low prediction errors and robust models.

\begin{figure*}%[ht]
\vskip 0.2in
\begin{center}
\centerline{\includegraphics[width=\textwidth]{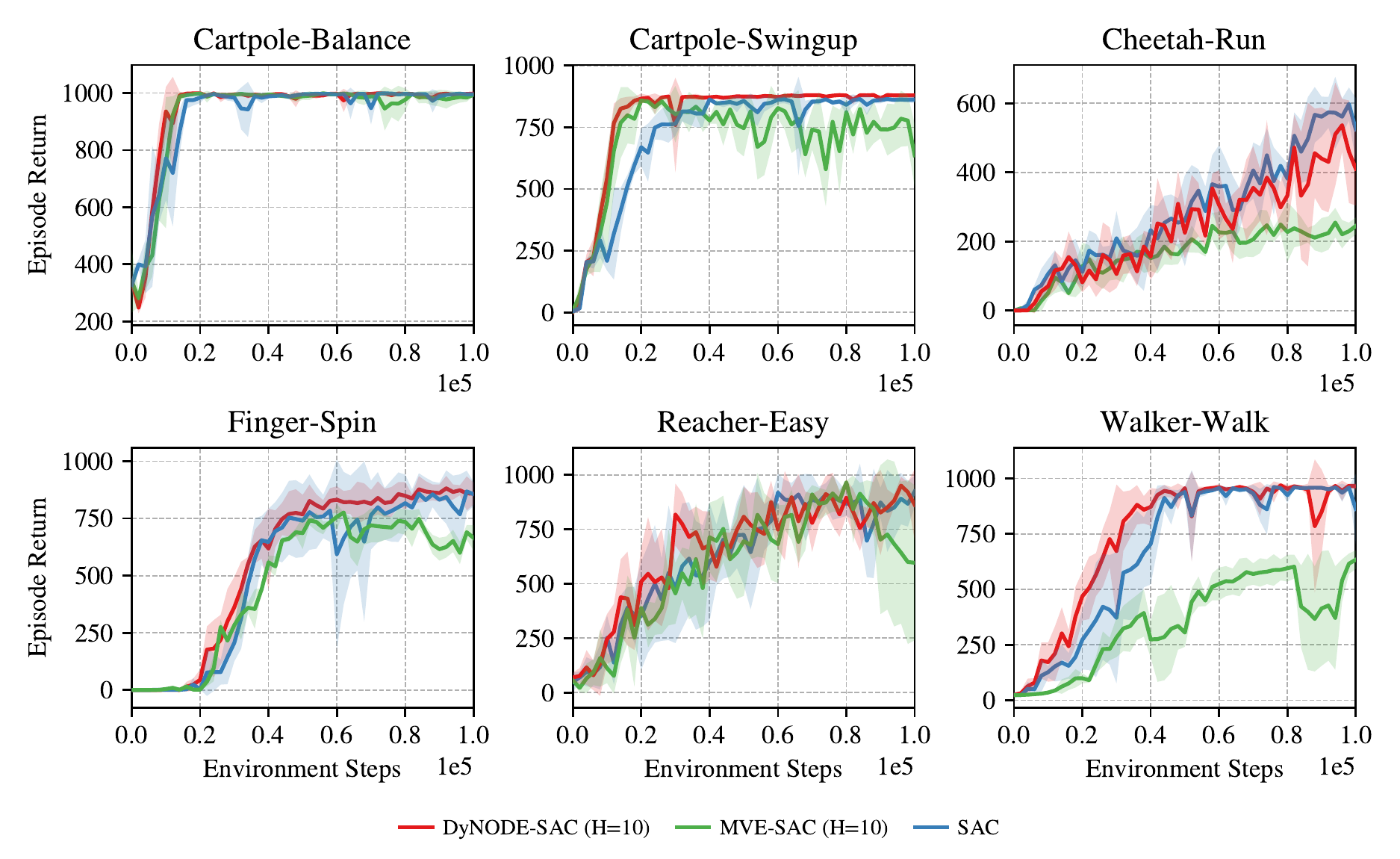}}
\caption{Performance of DyNODE coupled to SAC averaged across
5 random seeds relative to MVE-SAC and SAC
baselines. At 100k environment steps DyNODE
achieves better asymptotic performance while being also more data-efficient and stable that the MVE-SAC baseline.}
\label{figure_8}
\end{center}
\vskip -0.3in
\end{figure*}

\subsection{DyNODE with SAC for MBRL}
We now present the results of a practical instantiation of a model-based RL algorithm that uses the dynamics model presented in the previous sections. DyNODE is a general framework to learn dynamics model and can, in principle, be combined with any RL algorithm that uses the model either as an auxiliary task to improve sample-efficiency or to sample rollouts and plan. To demonstrate the benefits of our method in comparison with a neural network baseline, we couple DyNODE with the soft-actor-critic (SAC)~\citep{haarnoja2018soft} algorithm, and then use the model predictions to improve value estimation as in MVE ~\citep{feinberg2018model}. The value function is trained on both real and imaginary states via the TD(k) trick. It has been shown that this simple trick can improve sample efficiency even when using a deterministic neural network as the transition function~\citep{feinberg2018model}. However, for this idea to work, TD(k) requires accurate model predictions into the future, which is the main advantage of using the DyNODE framework.

In Fig.~\ref{figure_8} we show the learning curves for DyNODE coupled to SAC for 6 DeepMind Control Suite environments relative to MVE-SAC and SAC baselines (shadowed regions indicate standard errors). At 100k environment steps DyNODE achieves better asymptotic performance while being also more data-efficient and stable that the MVE-SAC baseline. More specifically, MVE-SAC improve performance in $2$ of $6$ tasks, while DyNODE-SAC improves performance in $5$ out of $6$ environments. These results also suggest that DyNODE can be a more effective approach to exploit a model for planning. In Cheetah-Run, both DyNODE-SAC and MVE-SAC fail to outperform the SAC baseline, indicating the presence of some instabilities in this environment, which was also pointed out by \cite{buckman2018sample}.

\begin{figure}%[ht]
\vskip 0.2in
\begin{center}
\centerline{\includegraphics[width=\columnwidth]{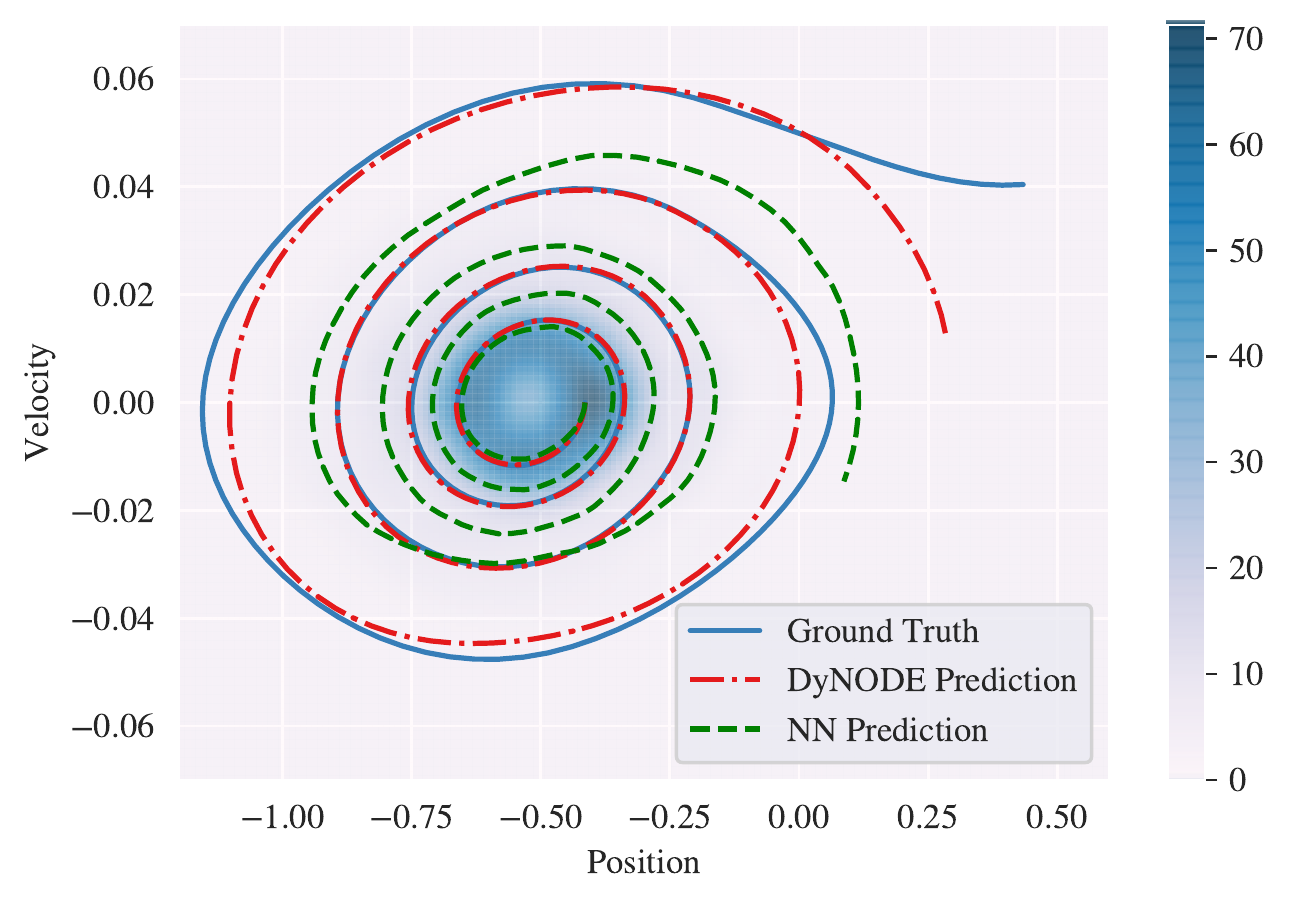}}
\caption{Phase-space dynamics evolution of near-optimal action sequences and the reconstruction of the optimal trajectory with the DyNODE-RK4 model and the neural network baseline for the MountainCar environment. The phase plot clearly shows how the momentum and position increase with time-steps resulting in a spiral trajectory. We also depict the density map of training data points (blue region in the center) used during training, corresponding to a single random rollout ($1000$ data points).}
\label{figure_3}
\end{center}
\vskip -0.2in
\end{figure}

\subsection{Trained Policy Trajectories}
Finally, in order to test the DyNODE model performance on data outside the training distribution, we use a trained policy to generate near-optimal action sequences for the MountainCar environment. As mentioned before, we chose this environment because the low dimensionality of its phase space allows for easy visualization and interpretation. 
In this phase-space dynamics, every phase-point represents the complete physical state (or description) of the system under consideration. We compared the DyNODE performance against the neural network baseline at predicting the trajectories generated by these sequences of actions. To further illustrate the ability of the DyNODE to generalize outside the training distribution, we attempted to reconstruct the optimal policy trajectory in phase space using both the DyNODE-RK4 model and the neural network baseline in the case where only a single random rollout (i.e. $1000$ data points) was used during training. Fig.~\ref{figure_3} presents a density map of the training points belonging to one random trajectories. The figure also shows that DyNODE can generalize well to unseen data and has a significantly better predictive performance then the neural network baseline.

\section{Conclusion}

We presented a new approach to train and learn dynamics models, where instead of using a canonical neural network, we have integrated control into a neural ordinary differential equations model in order to better capture the underlying dynamics that dictate the evolution of the environment. The results presented show that the DyNODE model outperforms the neural network baseline in both predictive performance and sample efficiency in the majority of the tested environments. In most cases, DyNODE trained with less training data surpasses other models that were trained with more data in terms of prediction ability, indicating that better generalization can be achieved with fewer training samples. Notice that, in this way, we are able to reduce the extrapolation error since the model mitigates the effect of compounding error as the length of the trajectory increases.

The experiments presented here were limited to using only simple, explicit methods for numerical integration, but still allowing DyNODE to outperform the neural network baseline. Moreover, extending DyNODE to a probabilistic or stochastic differential equation framework will allow us to propagates uncertainty state distributions through a NODE dynamics model while being better at generalization. Lastly, we want to remark that some other hybrid methods combining model-free and model-based RL approaches, such as the Dyna-style algorithms, could also benefit from having access to a more accurate model, like the one presented in this work.

\appendix

\subsubsection*{Acknowledgments}

This work was supported by the European Regional Development Fund and the Romanian Government through the Competitiveness Operational Programme 2014--2020, project ID P\_37\_679, MySMIS code 103319, contract no. 157/16.12.2016.

\bibliographystyle{cas-model2-names}

\bibliography{biblio}

\begin{thebibliography}{29}
\expandafter\ifx\csname natexlab\endcsname\relax\def\natexlab#1{#1}\fi
\providecommand{\url}[1]{\texttt{#1}}
\providecommand{\href}[2]{#2}
\providecommand{\path}[1]{#1}
\providecommand{\DOIprefix}{doi:}
\providecommand{\ArXivprefix}{arXiv:}
\providecommand{\URLprefix}{URL: }
\providecommand{\Pubmedprefix}{pmid:}
\providecommand{\doi}[1]{\href{http://dx.doi.org/#1}{\path{#1}}}
\providecommand{\Pubmed}[1]{\href{pmid:#1}{\path{#1}}}
\providecommand{\bibinfo}[2]{#2}
\ifx\xfnm\relax \def\xfnm[#1]{\unskip,\space#1}\fi
%Type = Misc
\bibitem[{Brockman et~al.(2016)Brockman, Cheung, Pettersson, Schneider,
  Schulman, Tang and Zaremba}]{open}
\bibinfo{author}{Brockman, G.}, \bibinfo{author}{Cheung, V.},
  \bibinfo{author}{Pettersson, L.}, \bibinfo{author}{Schneider, J.},
  \bibinfo{author}{Schulman, J.}, \bibinfo{author}{Tang, J.},
  \bibinfo{author}{Zaremba, W.}, \bibinfo{year}{2016}.
\newblock \bibinfo{title}{Openai gym}.
\newblock \URLprefix \url{http://arxiv.org/abs/1606.01540}. \bibinfo{note}{cite
  arxiv:1606.01540}.
%Type = Inproceedings
\bibitem[{Buckman et~al.(2018)Buckman, Hafner, Tucker, Brevdo and
  Lee}]{buckman2018sample}
\bibinfo{author}{Buckman, J.}, \bibinfo{author}{Hafner, D.},
  \bibinfo{author}{Tucker, G.}, \bibinfo{author}{Brevdo, E.},
  \bibinfo{author}{Lee, H.}, \bibinfo{year}{2018}.
\newblock \bibinfo{title}{Sample-efficient reinforcement learning with
  stochastic ensemble value expansion}, in: \bibinfo{booktitle}{Advances in
  Neural Information Processing Systems}, pp. \bibinfo{pages}{8224--8234}.
%Type = Incollection
\bibitem[{Chen et~al.(2018)Chen, Rubanova, Bettencourt and Duvenaud}]{NODE}
\bibinfo{author}{Chen, T.Q.}, \bibinfo{author}{Rubanova, Y.},
  \bibinfo{author}{Bettencourt, J.}, \bibinfo{author}{Duvenaud, D.K.},
  \bibinfo{year}{2018}.
\newblock \bibinfo{title}{Neural ordinary differential equations}, in:
  \bibinfo{editor}{Bengio, S.}, \bibinfo{editor}{Wallach, H.},
  \bibinfo{editor}{Larochelle, H.}, \bibinfo{editor}{Grauman, K.},
  \bibinfo{editor}{Cesa-Bianchi, N.}, \bibinfo{editor}{Garnett, R.} (Eds.),
  \bibinfo{booktitle}{Advances in Neural Information Processing Systems 31}.
  \bibinfo{publisher}{Curran Associates, Inc.}, pp.
  \bibinfo{pages}{6571--6583}.
%Type = Article
\bibitem[{Chua et~al.(2018)Chua, Calandra, McAllister and Levine}]{trials}
\bibinfo{author}{Chua, K.}, \bibinfo{author}{Calandra, R.},
  \bibinfo{author}{McAllister, R.}, \bibinfo{author}{Levine, S.},
  \bibinfo{year}{2018}.
\newblock \bibinfo{title}{Deep reinforcement learning in a handful of trials
  using probabilistic dynamics models}.
\newblock \bibinfo{journal}{CoRR} \bibinfo{volume}{abs/1805.12114}.
\newblock \href{http://arxiv.org/abs/1805.12114}{\tt arXiv:1805.12114}.
%Type = Article
\bibitem[{Clavera et~al.(2018)Clavera, Rothfuss, Schulman, Fujita, Asfour and
  Abbeel}]{MB-MPO}
\bibinfo{author}{Clavera, I.}, \bibinfo{author}{Rothfuss, J.},
  \bibinfo{author}{Schulman, J.}, \bibinfo{author}{Fujita, Y.},
  \bibinfo{author}{Asfour, T.}, \bibinfo{author}{Abbeel, P.},
  \bibinfo{year}{2018}.
\newblock \bibinfo{title}{Model-based reinforcement learning via meta-policy
  optimization}.
\newblock \bibinfo{journal}{CoRR} \bibinfo{volume}{abs/1809.05214}.
\newblock \href{http://arxiv.org/abs/1809.05214}{\tt arXiv:1809.05214}.
%Type = Inproceedings
\bibitem[{Deisenroth and Rasmussen(2011)}]{PILCO}
\bibinfo{author}{Deisenroth, M.}, \bibinfo{author}{Rasmussen, C.E.},
  \bibinfo{year}{2011}.
\newblock \bibinfo{title}{Pilco: A model-based and data-efficient approach to
  policy search}, in: \bibinfo{booktitle}{Proceedings of the 28th International
  Conference on machine learning (ICML-11)}, pp. \bibinfo{pages}{465--472}.
%Type = Article
\bibitem[{Feinberg et~al.(2018)Feinberg, Wan, Stoica, Jordan, Gonzalez and
  Levine}]{feinberg2018model}
\bibinfo{author}{Feinberg, V.}, \bibinfo{author}{Wan, A.},
  \bibinfo{author}{Stoica, I.}, \bibinfo{author}{Jordan, M.I.},
  \bibinfo{author}{Gonzalez, J.E.}, \bibinfo{author}{Levine, S.},
  \bibinfo{year}{2018}.
\newblock \bibinfo{title}{Model-based value estimation for efficient model-free
  reinforcement learning}.
\newblock \bibinfo{journal}{arXiv preprint arXiv:1803.00101} .
%Type = Inproceedings
\bibitem[{Gal et~al.(2016)Gal, McAllister and Rasmussen}]{gal2016improving}
\bibinfo{author}{Gal, Y.}, \bibinfo{author}{McAllister, R.},
  \bibinfo{author}{Rasmussen, C.E.}, \bibinfo{year}{2016}.
\newblock \bibinfo{title}{Improving pilco with bayesian neural network dynamics
  models}, in: \bibinfo{booktitle}{Data-Efficient Machine Learning workshop,
  ICML}, p.~\bibinfo{pages}{34}.
%Type = Article
\bibitem[{Haarnoja et~al.(2018)Haarnoja, Zhou, Abbeel and
  Levine}]{haarnoja2018soft}
\bibinfo{author}{Haarnoja, T.}, \bibinfo{author}{Zhou, A.},
  \bibinfo{author}{Abbeel, P.}, \bibinfo{author}{Levine, S.},
  \bibinfo{year}{2018}.
\newblock \bibinfo{title}{Soft actor-critic: Off-policy maximum entropy deep
  reinforcement learning with a stochastic actor}.
\newblock \bibinfo{journal}{arXiv preprint arXiv:1801.01290} .
%Type = Article
\bibitem[{Heess et~al.(2015)Heess, Wayne, Silver, Lillicrap, Tassa and
  Erez}]{SVG}
\bibinfo{author}{Heess, N.}, \bibinfo{author}{Wayne, G.},
  \bibinfo{author}{Silver, D.}, \bibinfo{author}{Lillicrap, T.P.},
  \bibinfo{author}{Tassa, Y.}, \bibinfo{author}{Erez, T.},
  \bibinfo{year}{2015}.
\newblock \bibinfo{title}{Learning continuous control policies by stochastic
  value gradients}.
\newblock \bibinfo{journal}{CoRR} \bibinfo{volume}{abs/1510.09142}.
\newblock \href{http://arxiv.org/abs/1510.09142}{\tt arXiv:1510.09142}.
%Type = Article
\bibitem[{{Khansari-Zadeh} and {Billard}(2011)}]{mixture}
\bibinfo{author}{{Khansari-Zadeh}, S.M.}, \bibinfo{author}{{Billard}, A.},
  \bibinfo{year}{2011}.
\newblock \bibinfo{title}{Learning stable nonlinear dynamical systems with
  gaussian mixture models}.
\newblock \bibinfo{journal}{IEEE Transactions on Robotics}
  \bibinfo{volume}{27}, \bibinfo{pages}{943--957}.
\newblock \DOIprefix\doi{10.1109/TRO.2011.2159412}.
%Type = Article
\bibitem[{Kurutach et~al.(2018)Kurutach, Clavera, Duan, Tamar and
  Abbeel}]{ME-TRPO}
\bibinfo{author}{Kurutach, T.}, \bibinfo{author}{Clavera, I.},
  \bibinfo{author}{Duan, Y.}, \bibinfo{author}{Tamar, A.},
  \bibinfo{author}{Abbeel, P.}, \bibinfo{year}{2018}.
\newblock \bibinfo{title}{Model-ensemble trust-region policy optimization}.
\newblock \bibinfo{journal}{CoRR} \bibinfo{volume}{abs/1802.10592}.
\newblock \href{http://arxiv.org/abs/1802.10592}{\tt arXiv:1802.10592}.
%Type = Incollection
\bibitem[{Levine and Abbeel(2014)}]{GPS1}
\bibinfo{author}{Levine, S.}, \bibinfo{author}{Abbeel, P.},
  \bibinfo{year}{2014}.
\newblock \bibinfo{title}{Learning neural network policies with guided policy
  search under unknown dynamics}, in: \bibinfo{editor}{Ghahramani, Z.},
  \bibinfo{editor}{Welling, M.}, \bibinfo{editor}{Cortes, C.},
  \bibinfo{editor}{Lawrence, N.D.}, \bibinfo{editor}{Weinberger, K.Q.} (Eds.),
  \bibinfo{booktitle}{Advances in Neural Information Processing Systems 27}.
  \bibinfo{publisher}{Curran Associates, Inc.}, pp.
  \bibinfo{pages}{1071--1079}.
%Type = Article
\bibitem[{Levine et~al.(2015a)Levine, Finn, Darrell and Abbeel}]{levi}
\bibinfo{author}{Levine, S.}, \bibinfo{author}{Finn, C.},
  \bibinfo{author}{Darrell, T.}, \bibinfo{author}{Abbeel, P.},
  \bibinfo{year}{2015}a.
\newblock \bibinfo{title}{End-to-end training of deep visuomotor policies}.
\newblock \bibinfo{journal}{CoRR} \bibinfo{volume}{abs/1504.00702}.
\newblock \href{http://arxiv.org/abs/1504.00702}{\tt arXiv:1504.00702}.
%Type = Article
\bibitem[{Levine et~al.(2015b)Levine, Wagener and Abbeel}]{GPS2}
\bibinfo{author}{Levine, S.}, \bibinfo{author}{Wagener, N.},
  \bibinfo{author}{Abbeel, P.}, \bibinfo{year}{2015}b.
\newblock \bibinfo{title}{Learning contact-rich manipulation skills with guided
  policy search}.
\newblock \bibinfo{journal}{CoRR} \bibinfo{volume}{abs/1501.05611}.
\newblock \href{http://arxiv.org/abs/1501.05611}{\tt arXiv:1501.05611}.
%Type = Article
\bibitem[{Lillicrap et~al.(2015)Lillicrap, Hunt, Pritzel, Heess, Erez, Tassa,
  Silver and Wierstra}]{Lili}
\bibinfo{author}{Lillicrap, T.P.}, \bibinfo{author}{Hunt, J.J.},
  \bibinfo{author}{Pritzel, A.}, \bibinfo{author}{Heess, N.M.O.},
  \bibinfo{author}{Erez, T.}, \bibinfo{author}{Tassa, Y.},
  \bibinfo{author}{Silver, D.}, \bibinfo{author}{Wierstra, D.},
  \bibinfo{year}{2015}.
\newblock \bibinfo{title}{Continuous control with deep reinforcement learning}.
\newblock \bibinfo{journal}{CoRR} \bibinfo{volume}{abs/1509.02971}.
%Type = Article
\bibitem[{Mnih et~al.(2015)Mnih, Kavukcuoglu, Silver, Rusu, Veness, Bellemare,
  Graves, Riedmiller, Fidjeland, Ostrovski, Petersen, Beattie, Sadik,
  Antonoglou, King, Kumaran, Wierstra, Legg and Hassabis}]{mnih}
\bibinfo{author}{Mnih, V.}, \bibinfo{author}{Kavukcuoglu, K.},
  \bibinfo{author}{Silver, D.}, \bibinfo{author}{Rusu, A.A.},
  \bibinfo{author}{Veness, J.}, \bibinfo{author}{Bellemare, M.G.},
  \bibinfo{author}{Graves, A.}, \bibinfo{author}{Riedmiller, M.},
  \bibinfo{author}{Fidjeland, A.K.}, \bibinfo{author}{Ostrovski, G.},
  \bibinfo{author}{Petersen, S.}, \bibinfo{author}{Beattie, C.},
  \bibinfo{author}{Sadik, A.}, \bibinfo{author}{Antonoglou, I.},
  \bibinfo{author}{King, H.}, \bibinfo{author}{Kumaran, D.},
  \bibinfo{author}{Wierstra, D.}, \bibinfo{author}{Legg, S.},
  \bibinfo{author}{Hassabis, D.}, \bibinfo{year}{2015}.
\newblock \bibinfo{title}{Human-level control through deep reinforcement
  learning}.
\newblock \bibinfo{journal}{Nature} \bibinfo{volume}{518},
  \bibinfo{pages}{529--533}.
%Type = Article
\bibitem[{Montgomery and Levine(2016)}]{GPS3}
\bibinfo{author}{Montgomery, W.}, \bibinfo{author}{Levine, S.},
  \bibinfo{year}{2016}.
\newblock \bibinfo{title}{Guided policy search as approximate mirror descent}.
\newblock \bibinfo{journal}{CoRR} \bibinfo{volume}{abs/1607.04614}.
\newblock \href{http://arxiv.org/abs/1607.04614}{\tt arXiv:1607.04614}.
%Type = Article
\bibitem[{Moore and Atkeson(1993)}]{moore}
\bibinfo{author}{Moore, A.W.}, \bibinfo{author}{Atkeson, C.G.},
  \bibinfo{year}{1993}.
\newblock \bibinfo{title}{Prioritized sweeping: Reinforcement learning with
  less data and less time}.
\newblock \bibinfo{journal}{Mach. Learn.} \bibinfo{volume}{13},
  \bibinfo{pages}{103–130}.
\newblock \URLprefix \url{https://doi.org/10.1023/A:1022635613229},
  \DOIprefix\doi{10.1023/A:1022635613229}.
%Type = Article
\bibitem[{Nagabandi et~al.(2017)Nagabandi, Kahn, Fearing and Levine}]{MBMF}
\bibinfo{author}{Nagabandi, A.}, \bibinfo{author}{Kahn, G.},
  \bibinfo{author}{Fearing, R.S.}, \bibinfo{author}{Levine, S.},
  \bibinfo{year}{2017}.
\newblock \bibinfo{title}{Neural network dynamics for model-based deep
  reinforcement learning with model-free fine-tuning}.
\newblock \bibinfo{journal}{CoRR} \bibinfo{volume}{abs/1708.02596}.
\newblock \href{http://arxiv.org/abs/1708.02596}{\tt arXiv:1708.02596}.
%Type = Article
\bibitem[{Plappert et~al.(2017)Plappert, Houthooft, Dhariwal, Sidor, Chen,
  Chen, Asfour, Abbeel and Andrychowicz}]{noise}
\bibinfo{author}{Plappert, M.}, \bibinfo{author}{Houthooft, R.},
  \bibinfo{author}{Dhariwal, P.}, \bibinfo{author}{Sidor, S.},
  \bibinfo{author}{Chen, R.Y.}, \bibinfo{author}{Chen, X.},
  \bibinfo{author}{Asfour, T.}, \bibinfo{author}{Abbeel, P.},
  \bibinfo{author}{Andrychowicz, M.}, \bibinfo{year}{2017}.
\newblock \bibinfo{title}{Parameter space noise for exploration}.
\newblock \bibinfo{journal}{CoRR} \bibinfo{volume}{abs/1706.01905}.
\newblock \URLprefix \url{http://arxiv.org/abs/1706.01905},
  \href{http://arxiv.org/abs/1706.01905}{\tt arXiv:1706.01905}.
%Type = Article
\bibitem[{Schulman et~al.(2017)Schulman, Wolski, Dhariwal, Radford and
  Klimov}]{schulman}
\bibinfo{author}{Schulman, J.}, \bibinfo{author}{Wolski, F.},
  \bibinfo{author}{Dhariwal, P.}, \bibinfo{author}{Radford, A.},
  \bibinfo{author}{Klimov, O.}, \bibinfo{year}{2017}.
\newblock \bibinfo{title}{Proximal policy optimization algorithms}.
\newblock \bibinfo{journal}{CoRR} \bibinfo{volume}{abs/1707.06347}.
\newblock \href{http://arxiv.org/abs/1707.06347}{\tt arXiv:1707.06347}.
%Type = Article
\bibitem[{Silver et~al.(2016)Silver, Huang, Maddison, Guez, Sifre, Van
  Den~Driessche, Schrittwieser, Antonoglou, Panneershelvam, Lanctot
  et~al.}]{silver2016mastering}
\bibinfo{author}{Silver, D.}, \bibinfo{author}{Huang, A.},
  \bibinfo{author}{Maddison, C.J.}, \bibinfo{author}{Guez, A.},
  \bibinfo{author}{Sifre, L.}, \bibinfo{author}{Van Den~Driessche, G.},
  \bibinfo{author}{Schrittwieser, J.}, \bibinfo{author}{Antonoglou, I.},
  \bibinfo{author}{Panneershelvam, V.}, \bibinfo{author}{Lanctot, M.}, et~al.,
  \bibinfo{year}{2016}.
\newblock \bibinfo{title}{Mastering the game of go with deep neural networks
  and tree search}.
\newblock \bibinfo{journal}{nature} \bibinfo{volume}{529},
  \bibinfo{pages}{484}.
%Type = Article
\bibitem[{Sutton(1991)}]{dyna}
\bibinfo{author}{Sutton, R.S.}, \bibinfo{year}{1991}.
\newblock \bibinfo{title}{Dyna, an integrated architecture for learning,
  planning, and reacting}.
\newblock \bibinfo{journal}{SIGART Bull.} \bibinfo{volume}{2},
  \bibinfo{pages}{160–163}.
\newblock \DOIprefix\doi{10.1145/122344.122377}.
%Type = Article
\bibitem[{Tassa et~al.(2018)Tassa, Doron, Muldal, Erez, Li, Casas, Budden,
  Abdolmaleki, Merel, Lefrancq et~al.}]{tassa2018deepmind}
\bibinfo{author}{Tassa, Y.}, \bibinfo{author}{Doron, Y.},
  \bibinfo{author}{Muldal, A.}, \bibinfo{author}{Erez, T.},
  \bibinfo{author}{Li, Y.}, \bibinfo{author}{Casas, D.d.L.},
  \bibinfo{author}{Budden, D.}, \bibinfo{author}{Abdolmaleki, A.},
  \bibinfo{author}{Merel, J.}, \bibinfo{author}{Lefrancq, A.}, et~al.,
  \bibinfo{year}{2018}.
\newblock \bibinfo{title}{Deepmind control suite}.
\newblock \bibinfo{journal}{arXiv preprint arXiv:1801.00690} .
%Type = Article
\bibitem[{Wang and Ba(2019)}]{want2019}
\bibinfo{author}{Wang, T.}, \bibinfo{author}{Ba, J.}, \bibinfo{year}{2019}.
\newblock \bibinfo{title}{Exploring model-based planning with policy networks}.
\newblock \bibinfo{journal}{arXiv preprint arXiv:1906.08649} .
%Type = Article
\bibitem[{Wang et~al.(2019)Wang, Bao, Clavera, Hoang, Wen, Langlois, Zhang,
  Zhang, Abbeel and Ba}]{benchmark}
\bibinfo{author}{Wang, T.}, \bibinfo{author}{Bao, X.},
  \bibinfo{author}{Clavera, I.}, \bibinfo{author}{Hoang, J.},
  \bibinfo{author}{Wen, Y.}, \bibinfo{author}{Langlois, E.},
  \bibinfo{author}{Zhang, S.}, \bibinfo{author}{Zhang, G.},
  \bibinfo{author}{Abbeel, P.}, \bibinfo{author}{Ba, J.}, \bibinfo{year}{2019}.
\newblock \bibinfo{title}{Benchmarking model-based reinforcement learning}.
\newblock \bibinfo{journal}{CoRR} \bibinfo{volume}{abs/1907.02057}.
\newblock \href{http://arxiv.org/abs/1907.02057}{\tt arXiv:1907.02057}.
%Type = Inproceedings
\bibitem[{Wiering and Schmidhuber(1998)}]{schmid}
\bibinfo{author}{Wiering, M.}, \bibinfo{author}{Schmidhuber, J.},
  \bibinfo{year}{1998}.
\newblock \bibinfo{title}{Efficient model-based exploration}, in:
  \bibinfo{booktitle}{PROCEEDINGS OF THE SIXTH INTERNATIONAL CONFERENCE ON
  SIMULATION OF ADAPTIVE BEHAVIOR: FROM ANIMALS TO ANIMATS 6},
  \bibinfo{publisher}{MIT Press/Bradford Books}. pp. \bibinfo{pages}{223--228}.
%Type = Article
\bibitem[{Xu et~al.(2018)Xu, Li, Tian, Darrell and Ma}]{SLBO}
\bibinfo{author}{Xu, H.}, \bibinfo{author}{Li, Y.}, \bibinfo{author}{Tian, Y.},
  \bibinfo{author}{Darrell, T.}, \bibinfo{author}{Ma, T.},
  \bibinfo{year}{2018}.
\newblock \bibinfo{title}{Algorithmic framework for model-based reinforcement
  learning with theoretical guarantees}.
\newblock \bibinfo{journal}{CoRR} \bibinfo{volume}{abs/1807.03858}.
\newblock \href{http://arxiv.org/abs/1807.03858}{\tt arXiv:1807.03858}.

\end{thebibliography}
\end{document}